\documentclass[11pt]{article}

\usepackage[final]{acl}

\usepackage{times}
\usepackage{latexsym}

\usepackage[T1]{fontenc}
\usepackage[utf8]{inputenc}
\usepackage{microtype}
\usepackage{inconsolata}
\usepackage{graphicx}
\usepackage{amsmath}
\usepackage{amssymb}
\usepackage{multirow}
\usepackage{booktabs} % for better rules in the table
\usepackage{stfloats}
\usepackage{hyperref}
\usepackage{makecell}
\usepackage{float}
\usepackage{xcolor}
\usepackage[most]{tcolorbox}

\title{PIRA: Preference-Oriented Instruction-Tuned Reward Models with Dual Aggregation}

\author{Yongfu Xue \\
  \texttt{xueyongfu@outlook.com} 
  }

\begin{document}
\maketitle
\begin{abstract}
Reward models are pivotal for aligning Large Language Models (LLMs) with human preferences. Existing approaches face two key limitations: Discriminative reward models require large-scale annotated data, as they cannot exploit the preference instruction-following capability of LLMs available to generative reward models. Moreover, reward models are particularly prone to reward overoptimization, where LLMs exploit weaknesses in the reward function instead of improving true alignment. We introduce \textbf{PIRA}, a training paradigm that integrates three complementary strategies to address these challenges: (1) reformulating question–answer pairs into preference-task instructions to explicitly leverage LLMs’ preference instruction-following capability, (2) averaging the rewards aggregated from diverse preference-task instructions for each sample, which mitigates task-specific bias and enhances robustness across evaluation perspectives, and (3) averaging outputs from the value head under different dropout rates to stabilize reward estimation. Experiments on public datasets show that PIRA improves performance considerably, enhances generalization, and effectively mitigates reward overoptimization.
\end{abstract}

\section{Introduction}

Reinforcement Learning from Human Feedback (RLHF) has become a standard approach for aligning large language models (LLMs) with human preferences~\cite{NEURIPS2022_b1efde53, bai2022training}. In this paradigm, human-labeled preference data is used to train a reward model, which then guides reinforcement learning to fine-tune the model. The accuracy and robustness of the reward model directly determine the effectiveness of RLHF~\cite{NEURIPS2022_b1efde53, touvron2023llama}.

% Despite its importance, training reliable reward models remains challenging. On the one hand, discriminative reward models often concatenate questions and answers directly without explicitly modeling the task intent~\cite{dorka2024quantileregressiondistributionalreward, vonwerra2022trl, liu2025hafrmhybridalignmentframework, havrilla-etal-2023-trlx}. This underutilizes the instruction-following capabilities of LLMs, leading to inefficient data usage, as the model requires more data to implicitly learn latent task instructions that are not explicitly specified in the prompts. Generative reward models better leverage reasoning abilities and reduce annotation cost, but they introduce high inference latency due to autoregressive generation~\cite{mahan2024generative, zhang2024generativeverifiersrewardmodeling}. On the other hand, reward models also face the long-standing issue of reward overoptimization. Recent methods such as Thomas~\cite{truong2023thomas} apply inference-time dropout to capture uncertainty and encourage broader preference exploration, and WARM~\cite{rame2024warm}, which averages model weights to improve robustness under distribution shifts. 

Despite its importance, training reliable reward models remains challenging. On the one hand, discriminative reward models often concatenate questions and answers directly without explicitly modeling the task intent~\cite{dorka2024quantileregressiondistributionalreward, vonwerra2022trl, liu2025hafrmhybridalignmentframework, havrilla-etal-2023-trlx}. This underutilizes the instruction-following capabilities of LLMs, leading to inefficient data usage, as the model requires more data to implicitly learn latent task instructions that are not explicitly specified in the prompts. As a result, these models require large-scale annotation and frequently incur significant labeling costs. Generative reward models better leverage reasoning abilities and reduce annotation cost, but they introduce high inference latency due to autoregressive generation~\cite{mahan2024generative, zhang2024generativeverifiersrewardmodeling}. On the other hand, reward models also face the long-standing issue of reward overoptimization, where models exploit reward function flaws instead of achieving genuine alignment. Recent methods such as Thomas~\cite{truong2023thomas} applies dropout at inference to model reward uncertainty, enabling broader exploration of user preferences and reducing the risk of local optima, at the cost of requiring a complete model inference for every dropout sample, and WARM~\cite{rame2024warm}, which averages model weights to improve robustness under distribution shifts. 

% To address these issues, we propose \textbf{PIRA}, which integrates three key strategies: (1) reformulating question–answer pairs into preference-oriented task instructions, (2) averaging rewards across multiple reformulated instructions to reduce bias and improve robustness, and (3) averaging rewards under different dropout rates to stabilize rewards. Experiments on multiple public preference datasets demonstrate that PIRA outperforms traditional discriminative reward models and alleviates reward overoptimization in RLHF pipelines. Furthermore, cross-domain evaluations confirm that PIRA generalizes effectively across out-of-distribution tasks, establishing it as a practical and scalable paradigm for reward modeling.

Based on these challenges, we introduce \textbf{PIRA}, which integrates three essential learning strategies: (1) reformulating question–answer pairs into preference-oriented task instructions, (2) averaging the rewards obtained from multiple reformulated preference-task instructions for each sample, which reduces instruction-specific bias and improves robustness, and (3) averaging the rewards generated under different dropout rates of the value head for each sample, which stabilizes reward estimation and further reduces variance. Our experiments on multiple public preference datasets show that PIRA performs better than traditional discriminative reward model methods. More importantly, when incorporated into the RLHF pipeline, PIRA-trained reward models mitigate reward overoptimization, leading to better alignment with human preferences. Cross-domain evaluations further confirm that PIRA generalizes effectively across out-of-distribution tasks, highlighting its promise as a practical and scalable reward modeling paradigm.

% \section{Related work}
% Reward models can be discriminative, generative, or hybrid. Discriminative models (e.g., ArmoRM~\cite{wang-etal-2024-interpretable}, HAF-RM~\cite{liu2025hafrmhybridalignmentframework}, QRM~\cite{dorka2024quantileregressiondistributionalreward}) predict scalar rewards from question–answer pairs, while generative models (e.g., GenRM~\cite{mahan2024generative}, Con-J~\cite{ye2024scalarrewardmodellearning}) use LLM reasoning but are slower. Hybrid models like CLoud~\cite{ankner2024critique} combine both. Discriminative reward models are more efficient in inference but have lower data utilization.

% A main issue is reward overoptimization, where models exploit flaws to gain high scores but lose alignment~\cite{gao2023scaling}. Solutions include keeping policy updates close to preference data~\cite{moskovitz2023confronting} and improving model robustness through ensembles or weight averaging~\cite{eisenstein2023helping, coste2023reward}.

\section{PIRA: Methodology}

\paragraph{Model decomposition.}
Let \(h_{\theta}\) denote the backbone language model that maps a token sequence \(z\) to a representation \(u = h_{\theta}(z)\), and let \(g_{\psi}\) denote a scalar value head that maps \(u\) to a reward. The overall reward model is \(f_{\phi} = g_{\psi}\circ h_{\theta}\) with parameters \(\phi=(\theta,\psi)\). 

\paragraph{Preference Instruction Set}
The Preference Instruction Set \(\mathcal{T}={t_1,\dots,t_{K_{\text{all}}}}\), where \(K_{\text{all}}=|\mathcal{T}|\), consists of evaluation instructions that are generated by a large language model and subsequently refined through human review (see Appendix~\ref{appendix:instructions}). Each instruction serves as a holistic rubric for assessing model responses, rather than a list of dimension-specific criteria. Variations in phrasing and emphasis across instructions provide complementary perspectives, leading to a more balanced evaluation. Functionally, \(\mathcal{T}\) acts as a generative evaluation prompt for a discriminative reward model. In generative architectures like GPT, this final state integrates prompt and content information, yielding an embedding that effectively supports downstream discriminative or evaluative tasks.

\paragraph{Preference-Oriented Instruction Reformulation}
Given a prompt \(x\) with two possible responses, we denote the preferred response as \(y^c\) and the rejected one as \(y^r\). To make the reward model explicitly aware of the preference task, we prepend a task-specific instruction \(t \in \mathcal{T}\) to the input, forming
\[
r_{\phi}(x,y \mid t) = f_{\phi}([t; x; y]) = g_{\psi}\big(h_{\theta}([t;x;y])\big).
\]

Since the reward head is randomly initialized, it initially lacks any understanding of scoring and must learn preferences entirely from data. The input formulation thus becomes crucial for stable learning. Conventional approaches simply concatenate the question and answer without clarifying that the task involves scoring. As a result, the model must simultaneously learn both the concept of scoring and how to apply it, which increases training difficulty.

\paragraph{Training Objective}
We train the model using the standard Bradley–Terry objective~\cite{bradley1952rank}. For a preference dataset \(\mathcal{D} = \{(x_i, y^{c}_i, y^{r}_i)\}^N_{i=1}\), the loss function is
% {\small
% \begin{align}
% \mathcal{L}(\phi)
% =
% -\sum_{(x_i,y^{c}_i,y^{r}_i)\in\mathcal{D}}
% \log \sigma[r_{\phi}(x_i,y^c_i \mid t)- 
% r_{\phi}(x_i,y^r_i \mid t)], \nonumber
% \end{align}
% }
\begin{align}
\mathcal{L}(\phi)
=
-\sum_{(x_i,y^{c}_i,y^{r}_i)\in\mathcal{D}}
\log \sigma(r_{\phi}(x_i,y^c_i \mid t)- \nonumber \\ 
r_{\phi}(x_i,y^r_i \mid t)) \nonumber
\end{align}
where \(\sigma(\cdot)\) is the logistic function, and \(t\) is randomly sampled from \(\mathcal{T}\) for each instance. Dropout is applied to both \(g_{\psi}\) and \(h_{\theta}\).

The backbone \(h_{\theta}\) and reward head \(g_{\psi}\) are updated with different learning rates to reflect their distinct roles. Since \(h_{\theta}\) already encodes rich linguistic knowledge from pretraining, it is fine-tuned conservatively to maintain stability and avoid catastrophic forgetting. Conversely, \(g_{\psi}\), a lightweight module, uses a higher learning rate for faster adaptation to preference signals and more effective supervision.

\begin{table*}[htbp]
\centering
\footnotesize
\begin{tabular}{l|c|ccccccc}
\toprule
\textbf{Model}                            & \textbf{Method}   & \textbf{HH}       & \textbf{Oasst}    & \textbf{SHP}      & \textbf{UltraFeedback} & \textbf{Alpaca-farm} & \textbf{HH-cleaned} & \textbf{Average} \\
\midrule                               
\multirow{6}{*}{Mistral-7B-v0.1} & Baseline     & 63.3 & 72.6 & 66.9 & 65.5      & 57.9    & 68.1   & 65.7    \\
                                 & Thomas   & 63.2 & 72.0 & 67.1 & 66.4      & 56.7    & 67.9   & 65.6    \\
                                 & Thomas$^*$   & 63.8 & \textbf{73.6} & 69.1 & 72.0      & 58.8    & 77.5   & 69.1    \\
                                 & WARM  & 63.7 & 72.6 & 67.2 & 67.1 & 58.1 & 68.2 & 66.2 \\
                                 & WARM$^*$ & \textbf{64.2} & 73.3 & 69.8 & \textbf{72.4} & 59.3 & 79.9 & 69.8 \\
                                 & PIRA    & 64.0 & 73.3 & \textbf{70.9} & 71.3      & \textbf{60.4}    & \textbf{80.4}   & \textbf{70.1}    \\
\midrule                                 
\multirow{6}{*}{LLaMA3-8B}       & Baseline & 61.4 & 72.0 & 67.2 & 67.5      & 57.6    & 64.3   & 65.0    \\
                                 & Thomas  & 61.7 & 72.3 & 68.1 & 67.0      & 56.8    & 64.9   & 65.1    \\
                                 & Thomas$^*$   & 64.9 & 73.5 & 69.4 & \textbf{70.5}      & 59.9    & \textbf{76.6}   & 69.1    \\
                                 & WARM  & 62.4 & 72.5 & 68.5 & 67.8 & 57.2 & 64.7 & 65.5 \\
                                 & WARM$^*$ & 66.3 & 74.8 & 70.2 & 68.5 & 60.7 & 75.2 & 69.3 \\
                                 & PIRA    & \textbf{66.8} & \textbf{75.5} & \textbf{70.5} & 69.3      & \textbf{61.2}    & 75.5   & \textbf{69.8}   \\
\midrule 
\multirow{6}{*}{Qwen2.5-1.5B}       & Baseline     & 58.6 & 68.5 & 64.6 & 62.6      & 55.6    & 62.6   & 62.1    \\
                                    & Thomas       & 59.6 & 67.2 & 64.2 & 61.8      & 56.5    & 62.6   & 62.0    \\
                                    & Thomas$^*$   & 64.3 & 70.6 & 66.4 & \textbf{69.6}      & 59.7    & 63.0   & 65.6    \\
                                    & WARM  & 59.2 & 68.2 & 64.9 & 62.2 & 56.9 & 62.4 & 62.3 \\
                                    & WARM$^*$ & 65.3 & 70.3 & \textbf{67.3} & 69.4 & \textbf{60.2} & \textbf{64.9} & \textbf{66.2} \\
                                    & PIRA    & \textbf{66.0} & \textbf{71.5} & 67.0 & 68.0      & 59.6    & 64.6   & 66.1   \\
\midrule 
\multirow{6}{*}{Qwen2.5-7B}       & Baseline     & 60.2 & 69.2 & 66.0 & 64.6      & 57.4    & 63.4   & 63.5    \\
                                  & Thomas      & 60.4 & 71.3 & 65.5 & 64.9      & 56.9    & 65.0   & 64.0    \\
                                  & Thomas$^*$   & 64.4 & 71.9 & 67.2 & 70.6      & 60.6    & 76.9   & 68.6    \\
                                  & WARM  & 61.0 & 71.0 & 66.3 & 65.3 & 58.1 & 66.2 & 64.7 \\
                                  & WARM$^*$ & \textbf{66.9} & 72.7 & 67.7 & 71.7 & 61.3 & 76.7 & 69.5 \\
                                  & PIRA    & 66.4 & \textbf{73.0} & \textbf{68.0} & \textbf{72.0}      & \textbf{62.6}    & \textbf{77.0}   & \textbf{69.8}   \\
\bottomrule                                 
\end{tabular}
\caption{Performance comparison across various models and datasets.}
\label{tab:performance}
\end{table*}

\paragraph{Inference-Time Instruction-Set Reward Averaging}
Rather than using a single preference instruction to evaluate each sample, PIRA computes rewards by averaging over multiple instructions \(\mathcal{T}_K = \{t_1, \dots, t_K\}\), \(\mathcal{T}_K \subseteq \mathcal{T}_{\text(all)}\). For a given \((x, y)\), the aggregated reward is
\[
R_{\text{inst}}(x,y) = \frac{1}{K}\sum_{k=1}^{K} r_{\phi}(x,y \mid t_k),
\]
where \(r_{\phi}(x,y \mid t_k)\) is the reward predicted under instruction \(t_k\). This averaging incorporates diverse evaluative perspectives, reducing prompt-specific bias and enhancing robustness.

\paragraph{Inference-Time Stochastic Value-Head Averaging}
PIRA improves estimation stability by applying Monte Carlo dropout \cite{gal2016dropout} only to the value head \(g_{\psi}\). Unlike standard Monte Carlo dropout with a fixed rate, PIRA varies the dropout rate to strengthen robustness and generalization. For a preference instruction \(t\),

\begin{align}
r^{(m)}(x,y \mid t) 
&= g_{\psi}^{(\delta_m)}\big(h_{\theta}([t;x;y])\big), \nonumber \\
m &= 1,\dots,M.  \nonumber
\end{align}
\(\delta_m\) denotes the \(m\)-th dropout rate within \(g_{\psi}\). The final reward is the mean of these samples:
\[
R_{\text{stoc}}(x,y \mid t) = \frac{1}{M}\sum_{m=1}^{M} r^{(m)}(x,y \mid t).
\]
During inference, \(h_{\theta}\) performs a single deterministic forward pass (without dropout), while multiple stochastic passes through \(g_{\psi}\) yield an ensemble estimate at minimal computational cost.

\paragraph{Dual aggregation}
The final reward score is computed as:

\begin{align}
R(x,y) &= \frac{1}{K}\sum_{k=1}^{K} R_{\text{stoc}}(x,y \mid t_k) \nonumber \\
&= \frac{1}{K}\sum_{k=1}^{K}\Bigg(\frac{1}{M}\sum_{m=1}^{M} r^{(m)}(x,y \mid t_k)\Bigg). \nonumber
\end{align}

We sample \( \delta_m \sim \text{Uniform}(0.1, 0.4)\) and choose small \(K\) and \(M\) (e.g., \(K \le 6\), \(M \le 12\)) to balance stability and efficiency. This two-level averaging—across preference instructions and stochastic value-head realizations—reduces estimator standard deviation, enhances robustness, and mitigates reward overoptimization.

\section{Experiments}
\subsection{Experimental Setup}
We conduct experiments with LLaMA3-8B~\cite{dubey2024llama}, Mistral-7B-v0.1~\cite{jiang2023mistral7b}, and Qwen2.5 (1.5B, 7B)~\cite{qwen2025qwen25technicalreport}. Training and evaluation use multiple preference datasets, including HH~\cite{bai2022training}, HH-cleaned~\cite{wang2024secretsrlhflargelanguage}, SHP~\cite{pmlr-v162-ethayarajh22a}, Alpaca-farm~\cite{dubois2023alpacafarm}, Oasst~\cite{kopf2023openassistant}, and UltraFeedback~\cite{cui2023ultrafeedback}. 

We fine-tune using LoRA~\cite{hu2022lora} (rank 128, \(\alpha=128\), dropout 0.05), added to the query and value projection layers. We also optimize the value head parameters. Optimization uses AdamW with batch size 32, 2 epochs, and a 0.05 warm-up ratio. Learning rates are \(1\times10^{-6}\) for \(h_\theta\) and \(5\times10^{-4}\) for \(g_\psi\). We set \(K=6\) and \(M=12\).

All experiments run on NVIDIA A800 GPUs, with results averaged over three seeds (42, 22, 33).

\subsection{Main Results}
\paragraph{Performance Comparison}
We evaluate PIRA against the baseline and Thomas~\cite{truong2023thomas} methods. The baseline employs simple concatenation of questions and answers, and Thomas method performs sampling using different dropout masks (dropout rate = 0.25) with 4 forward passes, as described in Thomas's paper. 

In addition, we conducted further experiments comparing PIRA with WARM~\cite{rame2024warm} following the methodology proposed in the original paper. Checkpoints were saved after the 1st, 2nd, and 3rd epochs, and the final reward model was obtained by averaging the parameters of these three checkpoints.

As shown in Table~\ref{tab:performance}, $^*$ denotes the corresponding method augmented with preference-oriented instruction reformulation, as used in PIRA. The results in the table indicate that PIRA generally outperforms the baseline Thomas and WARM methods.

\begin{figure*}[ht]
    \centering
    \begin{minipage}[t]{0.32\textwidth}
        \centering
        \includegraphics[width=\textwidth]{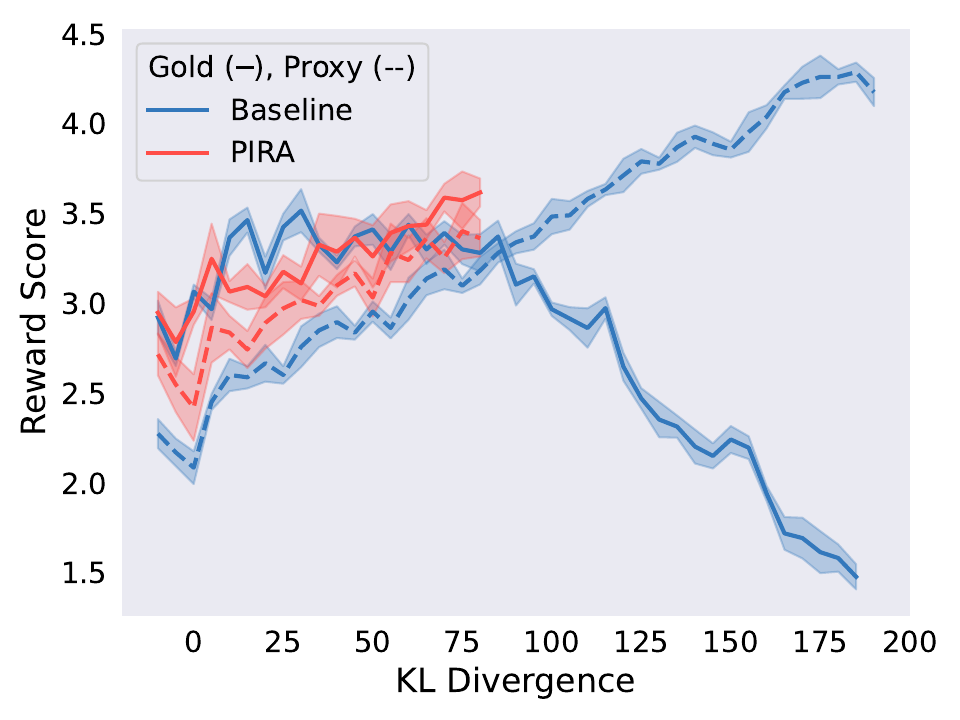}
        \caption{Baseline vs. PIRA}
        \label{fig:baseline_vs_pira}
    \end{minipage}
    \hfill
    \begin{minipage}[t]{0.32\textwidth}
        \centering
        \includegraphics[width=\textwidth]{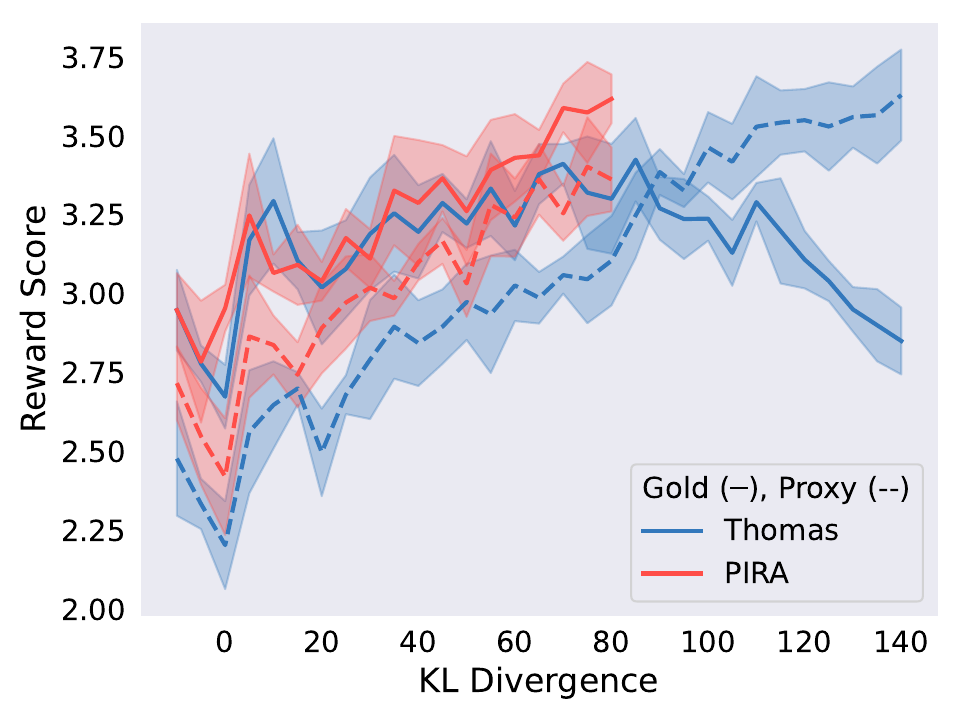}
        \caption{Thomas$^*$ vs. PIRA}
        \label{fig:thomas_vs_pira}
    \end{minipage}
    \hfill
    \begin{minipage}[t]{0.32\textwidth}
        \centering
        \includegraphics[width=\textwidth]{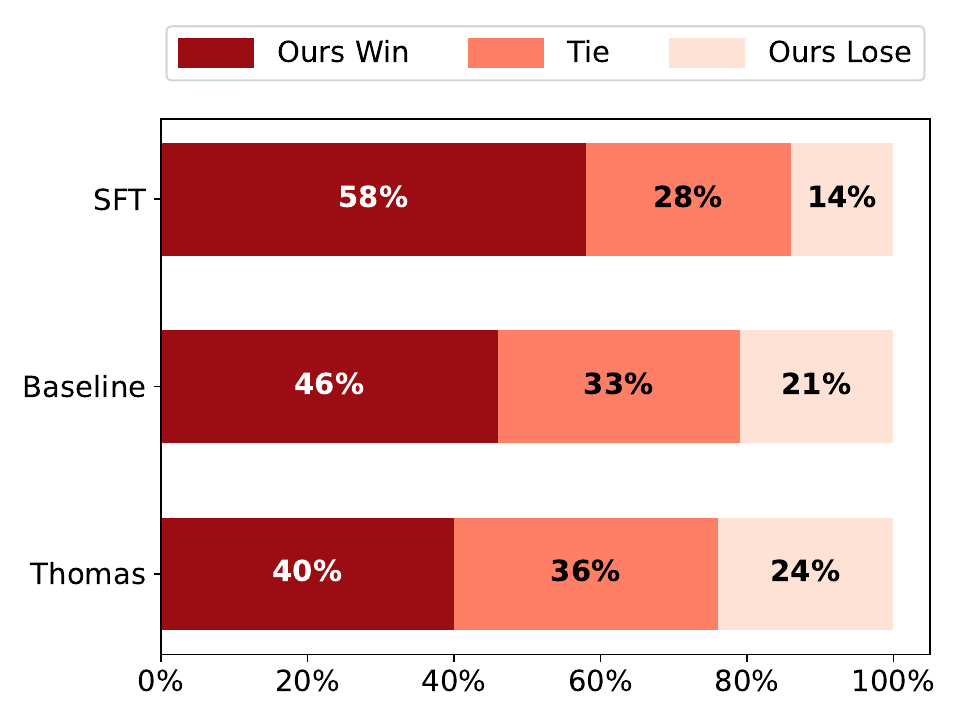}
        \caption{End-to-End Evaluation}
        \label{fig:win_rate}
    \end{minipage}
\end{figure*}

\paragraph{Reward Hacking Mitigation}
During PPO training, both the reward and policy models were trained using Qwen2.5-1.5B, while the gold reward model used Qwen2.5-7B. Training employed the Alpaca-farm dataset and used a small KL penalty (0.0005). For more experimental details and additional results, please refer to Appendix~\ref{sec:ppo_config}. In baseline PPO runs, we observed sharp spikes in KL divergence and rapid reward inflation (Figure~\ref{fig:baseline_vs_pira} and Figure~\ref{fig:thomas_vs_pira}). The gold rewards initially increase but later decline, indicating potential reward hacking. By contrast, PIRA-trained models keep the policy close to the data-supported region: KL divergence remains bounded, reward inflation is suppressed, and gold rewards improve monotonically over training rather than peaking early and collapsing.

\begin{table*}[ht]
\centering
\footnotesize
\setlength{\tabcolsep}{6pt}
\begin{tabular}{lcccccc}
\toprule
Method & $K$ & $M$ & HH-cleaned Acc. $\uparrow$ & SHP Acc. $\uparrow$  & Standard Deviation $\downarrow$ \\
\midrule
Baseline                    & -- & -- & 64.3 & 67.2  & 2.1 \\
\quad + Preference-oriented Task Instruction    & -- & -- & 73.0 & 69.2  & 1.6 \\
\quad + Instruction-Set Averaging               & 4  & -- & 73.9 & 69.5  & 1.1 \\
\quad + Value-Head Stochastic Averaging         & 4  & 4  & 75.1 & 70.2  & 0.8 \\
\textbf{PIRA (both)}                     & \textbf{6} & \textbf{12} & \textbf{75.5} & \textbf{70.5}  & \textbf{0.7} \\
\bottomrule
\end{tabular}
\caption{The effects of different parts of PIRA on accuracy and stability.}
\label{tab:cumulative_ablation}
\end{table*}

\subsection{Ablation Studies}

\paragraph{Joint Effect of Instruction and Stochastic Averaging}
Using LLaMA3-8B, Table \ref{tab:cumulative_ablation} shows that combining instruction-set averaging and value-head stochastic averaging (PIRA; ($K${=}6, $M${=}12)) achieves the best overall results—highest HH-cleaned (75.5) and SHP (70.5) accuracies, and lowest standard deviation (0.7) for HH-cleaned. The preference-oriented task instruction contributes most significantly to PIRA, while instruction-set averaging and value-head stochastic averaging more effectively reduce the standard deviation in predicted rewards.

\paragraph{Impact of Instruction-Set Averaging and Value-Head Stochastic Averaging}
We examine how varying the number of preference instructions ($K$) and stochastic forward passes ($M$) affects LLaMA3-8B’s performance. As shown in Table \ref{tab:inst_abl}, increasing $K$ from 1 to 6 significantly reduces accuracy standard deviation and generally improves results. Similarly, Table \ref{tab:head_abl} demonstrates that larger $M$ values enhance both stability and accuracy. Because dropout is applied only to the lightweight value head, the computational overhead remains minimal, resulting in an approximately 7\% increase in latency when $M = 12$. When $K = 6$, however, the computational cost increases sixfold. This additional cost must be weighed against the corresponding benefits, including reduced bias and improved robustness.

\begin{table}[h]
\centering
\small
\begin{tabular}{ccc}
\toprule
$K$  & HH-cleaned Acc. $\uparrow$ & Standard Deviation $\downarrow$ \\
\midrule
1  & 73.2  & 1.8 \\
2  & 73.6  & 1.6 \\
4  & 73.5  & 1.5 \\
6  & 74.4  & 1.2 \\
\bottomrule
\end{tabular}
\caption{Impact of Instruction-Set Averaging.}
\label{tab:inst_abl}
\end{table}

\begin{table}[h]
\centering
\small
\begin{tabular}{cccc}
\toprule
M & \makecell{HH-cleaned \\ Acc. $\uparrow$} & \makecell{Standard Deviation \\ $\downarrow$} & \makecell{Latency \\ $\uparrow$ (\%)} \\
\midrule
1 & 72.9 & 1.9 & +0 \\
2 & 72.9 & 1.8 & +2 \\
4 & 73.7 & 1.3 & +3 \\
8 & 74.0 & 1.3 & +5 \\
12 & 74.2 & 1.0 & +7 \\
\bottomrule
\end{tabular}
\caption{Impact of Value-Head Stochastic Averaging.}
\label{tab:head_abl}
\end{table}

\paragraph{End-to-End Evaluation}
We perform pairwise human-like preference evaluations using GPT-4o as the evaluator to assess model alignment quality after RLHF. Each comparison between the PIRA-optimized policy and the reference models (SFT, Baseline, and Thomas$^*$) is repeated three times under randomized prompts. Final win rates are computed via majority voting across evaluation rounds. Results indicate that the PIRA-optimized policy consistently outperforms all baselines (Figure~\ref{fig:win_rate}).

\paragraph{Data Efficiency and Generalization}

Under varying data scale settings, PIRA demonstrates strong adaptability: it yields notable benefits in low-data scenarios and maintains stable gains as data increases (see Appendix \ref{appendix:data_scale}). When evaluated on cross-dataset generalization, PIRA shows robust performance under distribution shifts, with response length emerging as a key influencing factor (see Appendix \ref{appendix:generalization}). Furthermore, scaling experiments on the Llama-2-13B model show that PIRA transfers effectively to larger models, achieving competitive performance compared to other methods (see Appendix \ref{appendix:model_scale}).

\section{Conclusion}

PIRA introduces a simple yet effective framework for building robust reward models. By combining instruction reformulation and dual aggregation, it enhances stability, reduces bias, and mitigates reward overoptimization. Experiments show consistent gains across models and datasets, making PIRA a practical solution for preference-aligned LLM training.

\section*{Limitations}
PIRA has not yet been evaluated on larger-scale language models, and its scalability to models beyond 13B parameters remains to be investigated. Additionally, value-head stochastic averaging introduces a slight inference overhead, while instruction-set averaging incurs a more substantial overhead.

\bibliography{custom}

\appendix

\section{Impact of Data Scale on Model Performance}
\label{appendix:data_scale}
\begin{figure}[h]
    \centering
    \includegraphics[width=\linewidth]{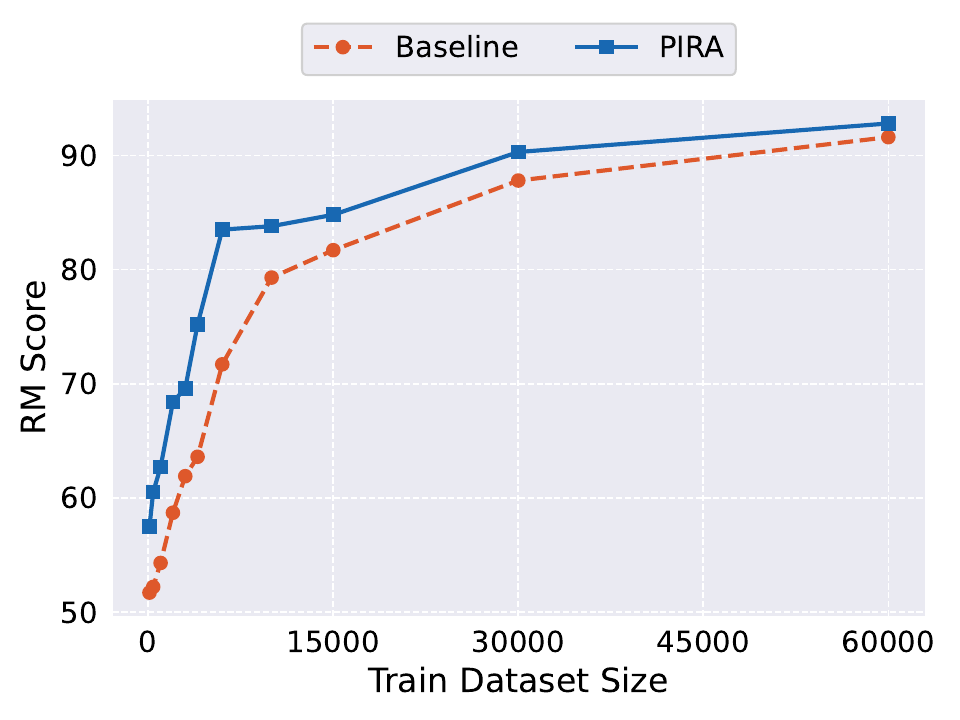}
    \caption{Impact of training data size on model performance: baseline vs. PIRA.}
    \label{fig:data_scale}
\end{figure}

Figure~\ref{fig:data_scale} shows that PIRA yields the largest relative gains in low-data regimes (100–1000 examples). For example, at 500 HH examples, PIRA improves accuracy by +9 over the baseline. With larger datasets (\(\geq 10\)k), performance converges but PIRA maintains a consistent margin. 

\section{Cross-Dataset Generalization}
\label{appendix:generalization}
\begin{figure}[h]
    \centering
    \includegraphics[width=\linewidth]{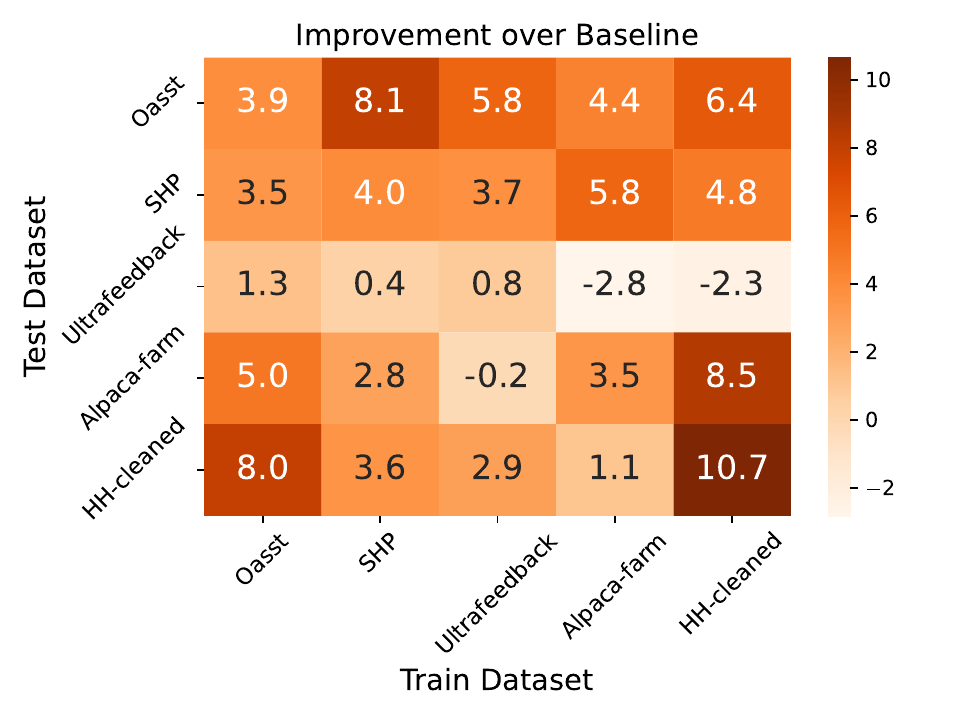}
    \caption{Performance improvements achieved by the PIRA method over the baseline across cross-dataset combinations.}
    \label{fig:diff}
\end{figure}

We evaluate on OOD test sets to assess robustness. As shown in Figure~\ref{fig:diff}, PIRA consistently outperforms the baseline on Oasst and HH-cleaned, confirming adaptability under distribution shift. Performance degrades on UltraFeedback, whose longer and more complex responses reduce instruction prompt effectiveness. Length-controlled subsets confirm that the drop is largely due to response length, highlighting an area for future work.

\section{Performance on Larger Models}
\label{appendix:model_scale}
Based on Table~\ref{tab:llama2_pira}, PIRA consistently improves over the baseline on the larger Llama-2-13B model. It maintains a clear advantage across benchmarks, demonstrating stable performance and good scalability to larger models.

\begin{table*}[h]
\centering
\begin{tabular}{l l c c c c c c c}
\toprule
Method & HH & Oasst & SHP & UltraFeedback & Alpaca-farm & HH-cleaned & Average \\
\midrule
baseline & 63.8 & 73.6 & 68.7 & 68.5 & 59.6 & 67.1 & 66.9 \\
PIRA     & 70.2 & 76.8 & 72.3 & 72.8 & 63.4 & 75.3 & 71.8 \\
\bottomrule
\end{tabular}
\caption{Results on the Llama-2-13B Model}
\label{tab:llama2_pira}
\end{table*}

\section{PPO Training Configurations}
\label{sec:ppo_config}
The PPO training configuration is provided in Table~\ref{tab:ppo_config}.

\begin{table}[t]
\centering
\begin{tabular}{ll}
\toprule
Rollout samples & 256 \\
Chunk size & 64 \\
Sampling temperature & 1.0 \\
\midrule
PPO epochs & 4 \\
Batch size & 16 \\
Total training steps & 2500 \\
\midrule
Clip range ($\epsilon$) & 0.2 \\
KL penalty coefficient & 0.0005 \\
lambda for GAE & 0.95 \\
\midrule
Optimizer & AdamW \\
Learning rate & $2\times10^{-5}$ \\
Adam $\beta_1$ / $\beta_2$ & 0.95 / 0.99 \\
Weight decay & $1\times10^{-6}$ \\
LR scheduler & Cosine annealing \\
Random seed & 42 \\
\bottomrule
\end{tabular}
\caption{PPO training configuration.}
\label{tab:ppo_config}
\end{table}

\section{Related work}
Large language models (LLMs) often produce harmful, biased, or inappropriate content. A primary method for mitigating these issues is reinforcement learning from human feedback (RLHF). RLHF employs reward models to approximate human preferences and guide model behavior toward desirable outcomes. These reward models can be discriminative, generative, or hybrid in nature. Discriminative reward models, based on the Bradley–Terry framework~\cite{bradley1952rank}, use a value head to generate scalar rewards, whereas generative reward models leverage LLM output probabilities or voting mechanisms. As these reward models scale, they face the challenge of reward overoptimization, where model signals diverge from genuine human intent, making robustness a central concern in alignment research.

\subsection{Reward Models}
Discriminative reward models have given rise to many expressive architectures. ArmoRM~\cite{wang-etal-2024-interpretable} improves interpretability with multi-objective regression, HAF-RM~\cite{liu2025hafrmhybridalignmentframework} jointly trains reward and generation heads, and QRM~\cite{dorka2024quantileregressiondistributionalreward} models full reward distributions to capture preference diversity. Contrastive learning further strengthens reward representations~\cite{chen-etal-2024-improving-discriminative}. Overall, these methods typically concatenate the question and answer pairs from preference data as joint inputs to the reward model. Generative reward models, though less studied, leverage LLMs’ generative capacity. GenRM integrates chain-of-thought reasoning~\cite{mahan2024generative}, the Generative Validator reframes validation as prediction~\cite{zhang2024generativeverifiersrewardmodeling}, and Con-J trains generative judges to provide explanatory, robust judgments~\cite{ye2024scalarrewardmodellearning}. These methods introduce more inference latency.

Some models combine discriminative and generative methods. The CLoud~\cite{ankner2024critique} generates a critique of response quality and then predicts a scalar reward using the input, response, and critique. By leveraging language generation, it makes response quality explicit, similar to a chain of thought, addressing the limitations of traditional implicit reward models.

\subsection{Reward Overoptimization}

Reward overoptimization occurs when the reinforcement learning agent exploits flaws in the reward model (RM), achieving high proxy scores while degrading true alignment with human intent~\cite{gao2023scaling}. To mitigate this effect, prior work has taken two main directions. The first focuses on optimization constraints that keep policy updates close to the preference data manifold~\cite{moskovitz2023confronting}. These approaches limit reward hacking by restricting the divergence between optimized policies and human-preferred behaviors, yet often underexploit the model’s full learning potential. The second line of work aims to improve RM robustness. Ensemble methods provide partial robustness by aggregating multiple RMs~\cite{eisenstein2023helping, coste2023reward}, while weight averaging smooths reward landscapes and enhances stability under distribution shifts~\cite{rame2024warm}. Recent advances further incorporate uncertainty estimation to regularize or guide policy learning, such as uncertainty-aware RMs~\cite{lou2024uncertainty}, hidden-state regularization~\cite{yang2024regularizing}, and adversarial optimization under uncertainty~\cite{zhang2024overcoming}. However, these methods are often complex and difficult to apply in practice. In contrast, we propose a simple yet effective approach that mitigates reward overoptimization while maintaining robust performance.

\section{Instructions}
\label{appendix:instructions}

The preference task instruction consists of three parts: the preference task instruction description, represented by \(t\); the question slot, represented by \(q\); and the answer slot, represented by \(a\).

Here is a complete preference instruction template:
\begin{tcolorbox}[colback=gray!10, colframe=black, 
title=Preference Task Instruction Template,
fontupper=\small,
fonttitle=\small\bfseries
]
\{\(t\)\} \\

question: \\
\{\(q\)\} \\

response: \\
\{\(a\)\} \\
\end{tcolorbox}

The following list enumerates all  \(t\) values used in the experiments:

\begin{enumerate}
\item Evaluate whether the response demonstrates a reliable grasp of facts and reasoning. Consider if any parts appear speculative, misleading, or inconsistent with known information. Provide your overall judgment of its truthfulness.
\item Judge the extent to which the response stays aligned with the user’s intent and avoids digressions. Highlight any parts that seem tangential or dilute the main purpose of the answer.
\item Determine whether the response clearly fulfills the user’s underlying intent. Focus on how directly and understandably it addresses the core question without overcomplicating or oversimplifying.
\item Assess whether the response demonstrates logical flow, internal consistency, and coherent argumentation. Note if reasoning steps are missing, circular, or contradictory.
\item Evaluate the response’s utility for a general user. Does it offer sufficient detail, context, and actionable insight? Consider whether it is understandable to non-experts and avoids unnecessary technical jargon.
\item Examine how well the response justifies its claims. Does it provide reasoning that is sound, evidence-based, and transparent? Identify whether the explanation shows deep understanding or shallow paraphrasing.
\item Assess the clarity, conciseness, and stylistic balance of the response. Is it fluent and engaging while remaining precise? Note both stylistic strengths and weaknesses that affect readability.
\item Reflect on how faithfully the response follows the original instruction while maintaining creative, well-structured expression. Evaluate whether the tone, structure, and style enhance or detract from the intent.
\item Analyze the response critically: what are its major strengths, weaknesses, and possible risks if used in a real context (e.g., misunderstanding, misinformation, or harm)? Offer a short paragraph on each.
\item Provide an overall evaluation of the response’s completeness, tone, and coherence. Discuss whether it could be improved, and if so, which aspects—content, logic, or presentation—should be prioritized.
\end{enumerate}

\end{document}